\documentclass{article}
\usepackage{spconf,graphicx}
\usepackage{color,soul}
\usepackage{adjustbox}
\usepackage{amsmath,amsthm,amssymb}
\title{Automatic Renal Segmentation in DCE-MRI using Convolutional Neural Networks}
\name{Marzieh Haghighi$^{\star \dagger}$ \qquad Simon K. Warfield $^{\star}$ \qquad Sila Kurugol$^{\star}$  }

\address{$^{\star}$Electrical and Computer Engineering Department, Northeastern University, Boston, MA, 02115\\
    $^{\dagger}$Radiology Department, Boston Children’s Hospital; and Harvard Medical School, Boston MA 02115}
\begin{document}
%\ninept
%
\maketitle
\begin{abstract}
Kidney function evaluation using dynamic contrast-enhanced MRI (DCE-MRI) images could help in diagnosis and treatment of kidney diseases of children. Automatic segmentation of renal parenchyma is an important step in this process. In this paper, we propose a time and memory efficient fully automated segmentation method which achieves high segmentation accuracy with running time in the order of seconds in both normal kidneys and kidneys with hydronephrosis. The proposed method is based on a cascaded application of two 3D convolutional neural networks that employs spatial and temporal information at the same time in order to learn the tasks of localization and segmentation of kidneys, respectively. Segmentation performance is evaluated on both normal and abnormal kidneys with varying levels of hydronephrosis. We achieved a mean dice coefficient of $91.4$ and $83.6$ for normal and abnormal kidneys of pediatric patients, respectively.
% We should put the results: We achieved a dice coefficient of xx and xx  over xx normal and xx abnormal kidneys respectively. 
\end{abstract}
\section{Introduction}
\label{sec:intro}
Hydronephrosis refers to the fluid-filled enlargement of the kidney as a result of obstruction in its output of urine. It is found in 2-7\% of all maternal ultrasound scans and 10\% of these children may have a significant urological problem. Delayed intervention in infants with severe hydronephrosis may lead to permanent loss of kidney function with the potential for lifelong complications associated with chronic renal insufficiencies. MRI can provide  clinically important markers of kidney function such as glomerular filtration rate (GFR) without exposing patient to ionizing radiation. Dynamic contrast-enhanced (DCE) MRI signal can be analyzed using pharmacokinetic (PK) models, and MR based GFR can be computed for determining whether a patient with persistent hydronephrosis will be referred for surgery or will receive conservative treatment. The PK models use  time intensity curves of the kidney parenchyma region to calculate both a single kidney GFR and a GFR map. Accurate segmentation of kidney parenchyma  is an important step to compute a robust and reliable GFR measure. Manual segmentation can take several hours. An accurate and robust technique for automated segmentation of kidney parenchyma (i.e. cortex and medulla) will reduce the burden on radiologist and accelerate the translation of MR based GFR technique into clinical practice.

DCE-MR image series include 3D volumetric images acquired at different time points after contrast injection.
The time intensity curves for different organs have different shapes. This temporal information can be used to discriminate kidney parenchyma from the other abdominal organs in the segmentation process.  Several automated segmentation techniques have been proposed for renal segmentation using this temporal information \cite{zollner2009assessment} and \cite{chevaillier2008functional}, however, for patients with diseased kidneys, these methods often fail. A software with user interface (CHOP-fMRU) is available for semi-automated segmentation and functional analysis of DCE-MR images \cite{khrichenko2010functional}, however, it requires several manual inputs from the user, such as drawing several initial boundary curves around the regions of interest. Recent studies has attempted to combine spatial and temporal information using a series of heuristic steps~\cite{yoruk2017automatic} ,\cite{yang2016renal}, some of which might fail in patients with pathological kidney and these approaches have relatively longer running times. Unlike previous segmentation techniques, deep learning algorithms process new test data very fast with running times on the order of seconds.
Another drawback of the previous methods is the usage of hand crafted features and thresholds which fail to perform well for patients with abnormal kidneys with enlarged pelvis and thinned parenchyma regions. In contrast, our proposed network learns a hierarchical representation of spatial and temporal features during the training process, which results in improved performance in both normal and abnormal kidneys.

%accurate assessment of renal function using DCE-MRI images has shown .... . Accurate automatic segmentation of kidneys can reduce GFR calculation time for new patients. DCE-MRI images are volumetric images which have temporal dimension. As the signal intensity time curves are different for different organs, these information can help discriminating kidney from ther organs in the segmentation process. Recent studies has attempted to combine spatial and temporal information using heuristic steps .... \cite{yoruk2017automatic} ,\cite{yang2016renal}. 
% polycystic
In this work, we propose a fully automated segmentation framework based on U-Net \cite{ronneberger2015u} architecture which was initially developed for 2D microscopy image segmentation. U-Net has the ability to capture local and global information for the image segmentation task and is able to generalize well from a small set of training samples. Variations of this network have also shown successful results for volumetric medical image segmentation \cite{cciccek20163d},\cite{milletari2016v},\cite{kamnitsas2017efficient}. We propose a 3D segmentation algorithm based on the 3D version of this architecture for automatic renal parenchyma segmentation in DCE-MR images. We will incorporate temporal features as the channel information for each voxel. However, applying the 3D U-Net for renal segmentation task has multiple challenges. First, each subject has a very large sized 4D data along with a large sized network parameter file, all of which can not be fitted into limited GPU memory and will therefore result in slow training process.  Moreover, the resulting segmentation usually needs a refining step such as dense conditional random fields \cite{christ2016automatic},\cite{kamnitsas2017efficient}, using auto-context \cite{salehi2017auto} or similar methods for reducing false positives which makes the algorithm more time inefficient.
Thus, we propose to divide our problem into two sub-problems that can be solved more efficiently in terms of time and memory when separated out: First, we apply a modified 3D U-Net on low resolution and augmented data for localizing the right and left kidneys; and second, we will apply U-Net on each extracted kidney region from the previous step for segmentation. Each of these sub-problems can be solved more quickly and need less memory compared to the naive approach. Our total test time is $<5$ seconds for each new patient.

% two drawbacks: 1) concatenation of time features in the channels makes the model parameters very large which results in large memory requirements and slow training time. 2) the resulting segmentation needs a refining step such as dense conditional random fields \cite{christ2016automatic},\cite{kamnitsas2017efficient}, using auto-context \cite{salehi2017auto} or similar methods for reducing false positives in imbalanced medical segmentation tasks with a large portion of data belong to the negative class. Thus, we propose to divide our problem into two sub-problems that can be solved more efficiently when separated out: First, we will use a simplified version of U-net for localizing the right and left kidneys; and second, we will apply U-Net on each extracted kidney region from the previous step for segmentation. Each of these sub-problems can be solved more quickly and need less memory compared to the naive approach. Our total test time is $<10$ seconds for each new patient.

%Considering these features, we propose to break our problem to two sub-problems which can be solved more efficiently comparing to the naive use of 3D U-Net and then refinement. We propose to: 1) use a simplified version of U-net for localization of right and left kidneys. 2) apply U-Net on each extracted kidney from previous step. Each of these sub problems can be solved faster and need less memory comparing to the naive approach. Total segmentation time is less than 60 seconds for each new patient. 

%%%%%%%%%%%%%%%%%%%%%%%%%%%%%%%%%%%%%%%%%%%%%%%%%%%%% Dataset
\section{Dataset}
\label{sec:dataset}
We use DCE-MRI images of 30 pediatric patients acquired at 3T for six minutes after injection of Gadavist using radial “stack-of-stars” 3D FLASH sequence (TR/TE/FA 3.56/1.39ms/12, 32 coronal slices, voxel size $1.25\times1.25\times3mm$). We retrospectively collected images from 30 patients with hydronephrosis who received MRI as part of their clinical protocol within the last 2 years. We also recruited 30 patients under a protocol approved by the Institutional
Review Board which specifically included recruiting subjects who receive contrast-enhanced MRI to undergo additional research imaging with DCE-MRI. We optimized acquisition protocol to achieve a mean temporal resolution of 3.3 sec for the arterial phase (2 minutes) and 13 sec for the remaining phases (4 minutes). 4D dynamic image series were reconstructed offline from raw data using a compressed sensing algorithm to improve temporal resolution and image quality, effectively reducing the streaking artifacts \cite{feng2014golden}. The age range of our pediatric patient group was between 2 months to 17 years. The field of view varied in patients with hydronephrosis according to the clinical protocol.

% I think we should omit this paragraph below
%Intra-subjects variability were exist in terms of different fields of view, image center, intensity range, kidney shapes (normal, abnormal), drift in sampling frequency (check with Sila). Data collection time was also varied due to subjects health conditions, bela bela....  . Number of time points for a 3D abdominal MRI image was at least x(check) and at most 156. 

%In the present study we used DCE-MRI images recorded from two groups of patients with normal and abnormal kidneys.

%%%%%%%%%%%%%%%%%%%%%%%%%%%%%%%%%%%%%%%%%%%%%%%%%%%%%% Methods
\section{Methods}
\label{sec:methods}
\begin{figure}[htb]

\begin{minipage}[b]{1.0\linewidth}
  \centering
  \centerline{\includegraphics[width=8.5cm]{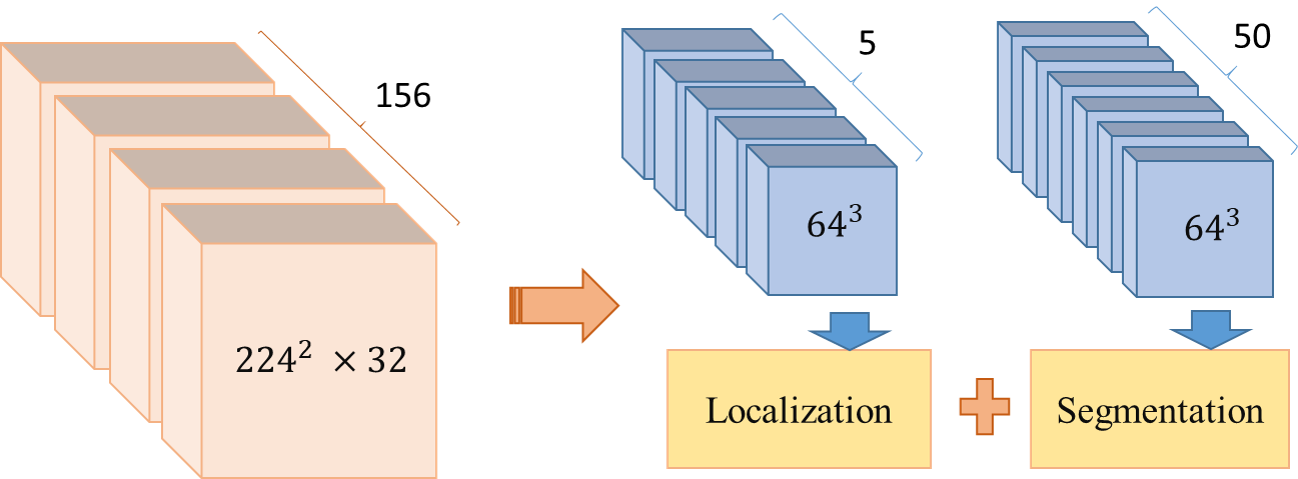}}
\end{minipage}
\caption{Automated segmentation steps: Original image segmentation problem is divided to: 1) localization based on low resolution data. 2) segmentation of each localized kidney.} %describe in detail}
\label{fig:overview}
\end{figure}

In this section, we describe a memory efficient renal segmentation framework which automatically segments kidneys given a 4D DCE-MRI as input. As described in Section~\ref{sec:dataset}, having a time dimension for each voxel creates a very large data tensor for each subject. On the other hand, U-Net architecture have shown to be successful in many MRI segmentation applications. However, the 3D version of this architecture is not memory efficient and requires many parameters to learn. Considering the nature of our 4D data, we need to reduce the data and network size in the context of fully convolutional network for the purpose of GPU processing.  Our proposed algorithm divide the problem into two sub-problems of localization and segmentation which can be solved more efficiently in terms of time and memory when separated out. Figure~\ref{fig:overview} summarizes the steps and the input data size of each step. We have used the fact that localization doesn't need high resolution image in space dimension and have used downsampled version of the image for localization. High resolution image inside the bounding box will be segmented in the second step. The preprocessing and the details of the networks used at each step are described in Section \ref{sec:loc} and \ref{sec:seg}.

% using In the first step, we preprocess the data in an specific manner and input it to the localization module. Both localization and segmentation steps are supervised learning processes based on Convolutional Neural Networks (CNN). The The subdivision of problem to two steps of localization and segmentation has mutple benefits including: 1)

% Four-dimensional DCE-MRI images were u ver had The data is preprocessesed  both two submodules ofInFour dimensional DCE-MRI images are first normalized to have equal range. Afterwards, we reduced time dimension information using PCA and kept first 5 dimensions with the highest variance. The reduced size data used for bounding box detection. The goal of this step was to reduce data size (mention gpu problem) and use it for rough segmentation. Rough segmentation was done using a modified u-net network which doesn't concatenate detail features in the first layers to the last layers and therefore is much smaller in terms of data and parameter space. After generating bounding boxes from first step, we use all the time features as the channel information to train a 3D unet architecture for accurate segmentation of kidneys in each bounding box. Detail of the architecture is summarized in the next subsections.

\vspace{-.5cm}
\subsection{Localization}
\label{sec:loc}
The 3D CNN network used for localization task is based on 3D U-Net and temporal features are mapped to channel information dimension of the network. U-net architecture learns the model with good generalization performance using a small number of training samples. However, considering our data variability described in Section \ref{sec:dataset}, data augmentation is necessary for the network to learn and generalize from a small number of  training samples. Given the large memory needed to load the training  data and network parameters, augmentation choice is very limited. The approach we used here was to reduce data size in dimensions where redundant information is present so that memory size will be reduced. This enabled us to use data augmentation in order to achieve an improved model fitting. To this end, we downsampled the data from $224\times 224\times32$ to $64\times 64\times 64$ to have sufficient resolution required for localization and  nearly isotropic resolution across different dimensions. We also reduced the time dimension using principle component analysis (PCA) and keep the first 5 dimensions with the highest variance. The 4D data of each subject was then augmented for various scales and feeded into the localization network. Augmentation details are given in Section~\ref{sec:res}. Localization network based on the U-net architecture consists of a contracting and an expansive path. Each layer in contracting path contains two $3\times3\times3$ convolution filters followed by a rectified linear unit (ReLu) and $2 \times 2 \times 2$ max pooling with strides of two for down-sampling. In the expansive path, each layer consists of a convolutional transpose of $2 \times 2 \times 2$ by strides of two in each dimension, followed by two $3 \times 3 \times 3$ convolutions each followed by a ReLu. Layers with equal resolution from contracting path are concatenated to their corresponding layers in the expansive path to add high-resolution features to the expansive path. Finally, a $1\times1\times1$ convolution reduces the number of output channels to the number of classes in the last layer. The input data to this network is a $64\times64\times64$ image with 5 channels. We used dropout layers after each maxpooling layer in the contracting path to reduce the chance of overfitting based on high resolution features in the first layers. Batch normalization was also used before the final $1\times1\times1$ convolution layer for having faster convergence and less overfitting. Input labels are forming three channels of foreground/background labels corresponding to each class.

% The input to the network is a 132 × 132 × 116 voxel tile of the image with 3
% channels. Our output in the final layer is 44×44×28 voxels in x, y, and z directions
% respectively. With a voxel size of 1.76×1.76×2.04µm3
% , the approximate receptive
% field becomes 155 × 155 × 180µm3
% for each voxel in the predicted segmentation.
% Thus, each output voxel has access to enough context to learn efficiently.
% We also introduce batch normalization (“BN”) before each ReLU. In [4],
% each batch is normalized during training with its mean and standard deviation
% and global statistics are updated using these values. This is followed by a layer
% to learn scale and bias explicitly. At test time, normalization is done via these
% computed global statistics and the learned scale and bias. However, we have
% a batch size of one and few samples. In such applications, using the current
% statistics also at test time works the best.

The network discriminates between three classes, namely right kidney, left kidney and background. However, there is an imbalanced distribution of samples in the kidney classes compared to the background class. We used a weighted cross entropy loss~\cite{christ2016automatic} in order to compensate for this imbalance and achieve accurate learning when training the fully convolutional network. Weighted cross entropy loss is given by  
\begin{equation}
L_{cross-entropy} = - \frac{1}{n}\sum_{i=1}^{N}w_{i}^{c}[\hat{p_{i}}\log p_{i}+ (1 - \hat{p_{i}}) \log(1 - p_{i})] 
\end{equation}
where $p_{i}$ is the probability of voxel $i$ belonging to the foreground in each output channel and $\hat{p_{i}}$ represents the true label in the corresponding input channel. We fix $w_{i}^{c}$ to be inversely proportional to the probability of voxel $i$ belonging to the foreground class. We used softmax with weighted cross-entropy loss for network output and true labels comparison. Cost minimization on 1000 epochs was performed using ADAM optimizer with learning rate of 0.0001. The training time for this network was approximately
one hour on a workstation with an NVIDIA Quadro 5000 GPU. % did not you use percival with GeForce GTX 1080.

\subsection{Segmentation}
\label{sec:seg}
We trained the second network, which performs the segmentation task, using the bounding boxes of the manually labeled kidneys in the training set. The kidneys were cropped and then fed into the second network for training. 
We resampled all cropped kidneys to a common spatial dimension of  $64 \times 64 \times 64$. We also interpolated and resampled the time intensity curves of each subject to a common temporal resolution and a common maximum acquisition time of 5 minutes. Fifty samples from 5 minutes acquisition were interpolated to ensure keeping the maximum variance of time intensity curves for different classes using minimum number of samples. The segmentation network used in this framework is the same as the localization network with the exception that the drop out layers were removed. The input data to this network is a $64\times64\times64$ image with 50 channels and input labels are two channels of foreground/background labels corresponding to each kidney/non-kidney class. We again used softmax along with weighted cross entropy loss to compare network output and true segmentation labels. Cost minimization on 500 epochs was performed using ADAM optimizer with learning rate of 0.0001. The training time for this network was approximately one hour on a workstation with an NVIDIA Quadro 5000 GPU.

% As it is also illustrated in Figure \ref{fig:networks}, it consists of contracting and expansive path. Each layer in contracting path contains x number of 3 x 3 x3 convolution filters followed by a rectified
% linear unit (ReLu) and $2 \times 2 \times 2$ max pooling with strides of two for down-sampling. In the expansive path, each layer consists of an convolutional transpose of $2 × 2 × 2$ by strides of two in each dimension, followed by two $3 \times 3 \times 3$ convolutions each followed by a ReLu. 

% Shortcut connections from layers of equal
% resolution in the analysis path provide the essential high-resolution features to the expansive path. Finally, a $1\times1\times1$ convolution reduces the number of output channels to the number of classes in the last layer. The input to the segmentation network is a $64 \times 64 \times 64$ volume with 156 channels which are the corresponding to each volume time features 

% \begin{figure}[htb]

% \begin{minipage}[b]{1.0\linewidth}
%   \centering
%   \centerline{\includegraphics[width=8cm]{image1.eps}}
% %  \vspace{2.0cm}
%   \centerline{(a) localization architecture based on u-net)}\medskip
% \end{minipage}
% %
% \begin{minipage}[b]{1\linewidth}
%   \centering
%   \centerline{\includegraphics[width=8cm]{image2.eps}}
% %  \vspace{1.5cm}
%   \centerline{(b) 3D unet)}\medskip
% \end{minipage}
% \caption{Localization and Segmentation network architecture based on U-net}
% \label{fig:networks}
% %
% \end{figure}

%%%%%%%%%%%%%%%%%%%%%%%%%%%%%%%%%%%%%%%%%%%%%%%%%%%%%

%%%%%%%%%%%%%%%%%%%%%%%%%%%%%%%%%%%%%%%%%%%%%%%%
\section{Experimental Results}
\label{sec:res}

\begin{figure}[htb]

\begin{minipage}[b]{1.0\linewidth}
  \centering
  \centerline{\includegraphics[width=8cm]{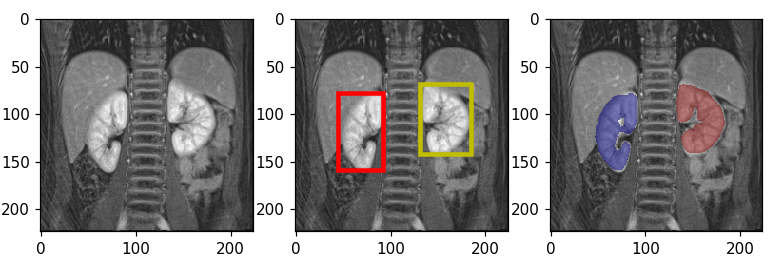}}
%  \vspace{2.0cm}
  \centerline{(a) Normal Kidney (F1-score: 94.63 $\%$)}\medskip
\end{minipage}
\vspace{-.5cm}
\begin{minipage}[b]{1\linewidth}
  \centering
  \centerline{\includegraphics[width=8cm]{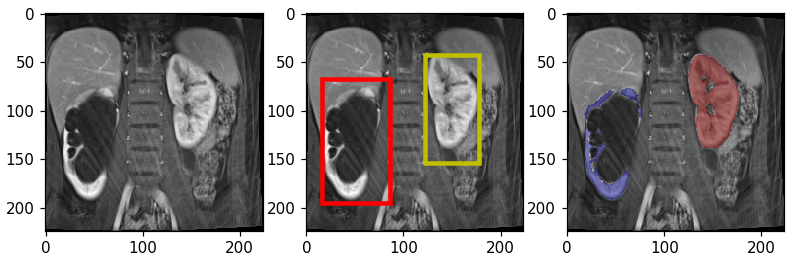}}
%  \vspace{1.5cm}
  \centerline{(b) Abnormal Kidney (F1-score: 88.27 $\%$)}\medskip
\end{minipage}
\caption{Kidney segmentation results:1) Localization using simplified 3D U-Net, 2) Segmentation of the cropped area using 3D U-Net}
\label{fig:res}
\end{figure}

To optimize the parameters of the proposed framework for automated segmentation of normal and abnormal kidneys, we performed cross validation experiments on 24 subjects (10 with normal and 14 with abnormal kidneys). We used precision, recall, dice coefficient (DSC) or F1-score and volumetric estimation error (VEE) for evaluating the algorithm segmentation performance. F1-score, which is the harmonic average of precision and recall, reports the accuracy of the overlap between the predicted and true manual segmentation. We also report the performance of the model, trained using 24 subjects, and tested on 12 kidneys from 6 previously unseen subjects (3 patients with normal and 3 patients with pathological kidneys) that were not included in the training process. As explained in section \ref{sec:methods}, we train each of the localization and the segmentation networks independently using the training data and the manual segmentation masks. Segmentation results are shown in Figure \ref{fig:res} for one normal and one abnormal kidney example from the test set. Middle figure in each row is showing the result of bounding box detection. Predicted output consisted of three classes; right kidney, left kidney and background. After extracting three classes from initial segmentation masks and forming the bounding boxes, each class was scaled to $64 \times 64 \times 64$ volumes and the original time dimension was resampled, interpolated and added to the data as the channel information. Finally, the segmentation classifies each voxel in the high resolution image into kidney or non-kidney class. Third figure in each row is showing the result of segmentation and re-positioning each kidney back into the detected bounding box. The resulted average performance measures for final unseen test cases are reported in Table.~\ref{tab:perf}. Mean F1-scores for three patients with normal and three with abnormal kidneys were $91.4$ and $83.6$ respectively.

\vspace{-.5cm}
\begin{table}[h!]
    \centering
    \caption{Renal segmentation performance.}
    \label{tab:kld}
       %\scalebox{0.8}[0.8] 
    \begin{adjustbox}{max width=6cm}
        \begin{tabular}{|l||*{3}{r|}}
        \hline
        Session & \multicolumn{2}{c|}{{\bf Test }(mean $\pm$ sd)}\\
        \hline
        Kidney &  Normal  & Abnormal  \\
        \hline  
        Precision&94.76$\pm$0.047	&90.56$\pm$0.027\\
%        Recall&87.73$\pm$0.027&xxx$\pm$x\\
        F1-score&91.43$\pm$0.034&83.65$\pm$0.033\\
        VEE&12.87$\pm$2.4 mL	&19.52$\pm$3.2 mL\\\hline
        Time &\multicolumn{2}{c|}{$\sim$ 3s}\\\hline
        \end{tabular}
    \end{adjustbox}% 
    \label{tab:perf}
    \vspace{-.5cm}
\end{table}
\section{Conclusion}
\label{sec:conc}
In this work we proposed a time and memory efficient fully-automated framework for segmentation of renal parenchyma using DCE-MRI data. The proposed learning based framework consists of two cascaded CNNs for localization and segmentation of kidneys. The proposed fully automated algorithm performed well in both normal and abnormal kidneys.

\section{ACKNOWLEDGEMENTS}
\label{sec:ack}
This work was supported by the Society of Pediatric Radiology Young Investigator Grant.
% List and number all bibliographical references at the end of the paper.  The references can be numbered in alphabetic order or in order of appearance in the document.  When referring to them in the text, type the corresponding reference number in square brackets as shown at the end of this sentence \cite{C2}.

\bibliographystyle{IEEEbib}
\bibliography{strings,refs}

\end{document}